\documentclass{article}
\usepackage{iclr2027_conference,times}
\iclrfinalcopy

\usepackage{amsmath,amsfonts,bm}

\def\eqref#1{equation~\ref{#1}}

\def\1{\bm{1}}

\DeclareMathAlphabet{\mathsfit}{\encodingdefault}{\sfdefault}{m}{sl}
\SetMathAlphabet{\mathsfit}{bold}{\encodingdefault}{\sfdefault}{bx}{n}

\usepackage{hyperref}
\usepackage{url}
\usepackage{graphicx}
\usepackage{xcolor}
\usepackage{amsmath}
\usepackage{amssymb}
\usepackage{booktabs}
\usepackage{colortbl}
\usepackage{tabularx}
\usepackage{float}
\usepackage{wrapfig}
\usepackage{placeins}
\usepackage{needspace}
\definecolor{paperLink}{HTML}{1F4E79}
\hypersetup{colorlinks=true,linkcolor=paperLink,citecolor=paperLink,urlcolor=paperLink,pdfauthor={Chun Gu, Xiaofei Wei, Yiming Chen, Juntong Zhang, Li Zhang},pdftitle={X-World: Transferring Video Priors to Variable-View Robot Control}}

\hypersetup{%
  colorlinks=true,
  citecolor=blue,
  linkcolor=red,
  urlcolor=magenta,
  pdfborder={0 0 0}%
}

\setcitestyle{authoryear,round,semicolon,aysep={,},yysep={,}}
\AtBeginDocument{\let\cite\citep}

\makeatletter
\def\paper@figuretype{figure}
\g@addto@macro\@floatboxreset{%
  \ifx\@captype\paper@figuretype
    \vspace*{-14pt}%
  \else
    \vspace*{-8pt}%
  \fi
}
\makeatother

\makeatletter
\def\paper@tabletype{table}
\long\def\@makecaption#1#2{%
  \vskip\abovecaptionskip
  \ifx\@captype\paper@tabletype
    \small\noindent\textbf{#1:}~#2\strut\par
    \vskip4pt
  \else
    \small\raggedright\noindent\textbf{#1:}~#2\par
    \vskip\belowcaptionskip
  \fi
}
\makeatother

\newcommand{\method}{\textit{X-World}}
\newcommand{\mainsectionref}[1]{Sec.~\ref{#1}}
\newcommand{\mainfigureref}[1]{Fig.~\ref{#1}}
\providecommand{\keywords}[1]{}

\definecolor{rankOne}{HTML}{B57FA6}
\definecolor{rankTwo}{HTML}{8A8FC3}
\newcommand{\first}[1]{\cellcolor{rankOne!22}\textbf{#1}}
\newcommand{\second}[1]{\cellcolor{rankTwo!22}\underline{#1}}

\title{X-World: Transferring Video Priors to Variable-View Robot Control}
\author{\parbox[t]{\dimexpr\textwidth-2\tabcolsep\relax}{\centering
  \normalsize
  Chun Gu$^{1,2,3}$, Xiaofei Wei$^{1,2,3}$, Yiming Chen$^{2,3}$,
  Juntong Zhang$^{2,3}$, Li Zhang$^{1,2,3\dagger}$\\[0.4em]
  \small\normalfont
  $^{1}$School of Data Science, Fudan University
  $^{2}$Shanghai Innovation Institute\quad $^{3}${\hypersetup{urlcolor=black}\href{https://github.com/LogosIntelligenceLab}{Logos Intelligence Lab}}\\[0.8em]
  \normalfont\href{https://logosintelligencelab.github.io/X-World/}{\texttt{logosintelligencelab.github.io/X-World/}}
}}

\begin{document}
\raggedbottom

\maketitle
\lhead{Preprint}
{\renewcommand{\thefootnote}{\fnsymbol{footnote}}%
\footnotetext[2]{Corresponding author (\href{mailto:lizhangfd@fudan.edu.cn}{\nolinkurl{lizhangfd@fudan.edu.cn}}).}}

\begin{abstract}
Video diffusion models provide a source of interaction representations for robot control, but transferring these representations requires reconciling variable camera inputs with heterogeneous action interfaces and closed-loop execution.
We present \method{}, a framework that transfers a video prior through a fixed-length token sequence to a shared action expert.
A residual cross-view attention branch extends a pretrained video backbone to one, two, or three views while retaining its original self-attention path.
A layer-aligned Q-Former aggregates backbone features, and dataset-specific modules map between native state and action interfaces and the shared action expert.
Training separates video adaptation from action-only pretraining by freezing the video backbone in the latter stage, followed by joint target-domain post-training.
The policy extracts video features with a single backbone pass and uses asynchronous action chunking for execution.
On LIBERO, \method{} reaches 98.5\% mean success; its LIBERO-trained checkpoint obtains 76.5\% on LIBERO-Plus without further adaptation.
The seven-shift breakdown reveals gains in robot, language, lighting, and layout perturbations, alongside weaknesses under camera, background, and noise shifts relative to OpenVLA-OFT.
On six dual-arm real-world tasks, it achieves 81.7\% mean success over 20 trials per task.
Removing video adaptation or cross-view fusion produces the largest observed success drops among the tested removals.
\end{abstract}

\section{Introduction}
\label{sec:intro}

Video data and robot action data offer different forms of supervision.
Videos capture object motion, contact, and task progression across many scenes, whereas action-labeled trajectories specify how a particular robot produces those transitions.
Vision-language-action policies~\cite{brohan2022rt,brohan2023rt,kim2024openvla,team2024octo,bjorck2025gr00t,black2024pi0,pi05} learn the latter mapping effectively, but collecting action supervision across robots and camera setups remains expensive.
Video diffusion models offer a complementary source of representations~\cite{ye2024latent,du2023learning,hu2024video}.
The question we study is how to transfer these representations to a policy whose observations and executable actions vary across datasets.

This transfer has two interfaces.
At the visual interface, datasets provide different numbers of cameras and different viewpoints.
A single-view video backbone must gain access to information across views without discarding its pretrained computation.
At the control interface, video hidden states contain many spatiotemporal tokens, while the policy must generate a short action chunk in the robot's native coordinates.
Exposing a changing visual context directly to an action expert couples these two forms of heterogeneity.
We instead use a fixed-length token sequence between the video backbone and the action expert, so camera processing and action decoding can be adapted separately.

\method{} instantiates this design with a pretrained LTX-Video backbone~\cite{hacohen2024ltx}, a layer-aligned Q-Former, and a flow-matching action expert.
The video backbone retains its pretrained self-attention and adds cross-view attention through a zero-initialized residual branch.
View embeddings, randomized view order, and masked replication support training on one-, two-, and three-view datasets.
The Q-Former aggregates features across backbone layers into the token sequence consumed by the action expert.
Following the use of dataset-specific interfaces in cross-embodiment learning~\cite{xvla}, lightweight projections map native state and action representations to and from the shared action backbone.

Training first adapts the video backbone on robot and human videos, then freezes it while learning the Q-Former and action expert from heterogeneous action-labeled trajectories.
Target-domain post-training updates the model with video and action supervision.
At deployment, one backbone pass extracts features from observed image latents and noisy future slots; the policy does not generate and decode a future video before acting.
Asynchronous real-time chunking~\cite{black2025real,black2025training} overlaps policy queries with execution.
On a single H100, the reported cached model response is 46\,ms; the deployed refresh, including communication and processing, is 91\,ms, while queued actions execute at 30\,Hz.

Our evaluation asks three questions: whether the transferred representation supports manipulation, which parts of the transfer account for the observed performance, and where robustness breaks down.
The policy reaches 98.5\% mean success on LIBERO~\cite{liu2023libero} and 76.5\% on zero-shot LIBERO-Plus~\cite{fei2025liberoplus}.
Because LIBERO is nearly saturated, we examine the full distribution-shift profile and the training-stage interaction rather than relying on its average ranking.
On six dual-arm UR5e tasks, \method{} achieves 81.7\% mean success, compared with 58.3\% for X-VLA, the strongest baseline in this real-world comparison.
Removing cross-view attention reduces LIBERO success to 83.1\% and Fold Towel success from 12/20 to 0/20, providing evidence beyond the saturated benchmark average.

This paper makes three main contributions:
\textbf{(i)} We introduce \textbf{\method{}}, a framework for variable-view robot control that adapts a pretrained video backbone through masked residual cross-view fusion while retaining its original self-attention path;
\textbf{(ii)} We build a video-to-action interface that combines a layer-aligned Q-Former with dataset-specific state and action projections, using fixed-length tokens and separating video adaptation from action-only pretraining;
and \textbf{(iii)} We analyze the benefits and limits of this transfer through training-stage ablations, seven distribution shifts, and dual-arm experiments, with separate measurements of model and system latency.

\section{Related work}
\label{sec:related}

\paragraph{From video models to robot actions.}
Video-action models differ in how they expose visual dynamics to control.
One approach generates future observations and maps predicted transitions to actions through inverse dynamics~\cite{du2023learning,du2023video,wen2024vidman}.
Another uses internal video representations as features for action generation~\cite{hu2024video,yang2025covar}.
Joint video-action architectures couple the two modalities within a transformer~\cite{bi2025motus,liao2025genie,ye2026world,li2026causal,pai2025mimic}, while Cosmos Policy~\cite{kim2026cosmos} encodes actions within a video generation formulation.
These works establish video-to-control transfer as an existing direction.
\method{} studies a particular interface within that direction: layer-aligned aggregation into a fixed-length token sequence, combined with a video backbone adapted to variable-view inputs.
At control time, it uses hidden states from one backbone pass rather than a decoded future rollout.

\paragraph{Multi-view adaptation.}
Genie-Envisioner~\cite{liao2025genie} builds multi-view video modeling on LTX-Video and provides a close reference for transferring video features to actions.
Our architectural distinction is to retain per-view pretrained self-attention and append cross-view fusion as a residual branch.
The added branch contributes zero at initialization while the original per-view computation is retained.
We evaluate residual fusion and view identity separately, and distinguish missing-view evaluation from robustness to camera-pose perturbations.

\paragraph{Heterogeneous action learning and continuous execution.}
Open-X-Embodiment~\cite{o2024open} and other multi-platform datasets~\cite{bu2025agibot,wu2025robocoin,jiang2025galaxea,wu2024robomind} enable training across robot interfaces.
Alignment layers, modular projections, and embodiment representations address differences in state and action conventions~\cite{doshi2024scaling,wang2024scaling,zheng2025universal,xvla}.
Other methods address observation differences through feature alignment~\cite{bian2025vla}, novel-view synthesis, or embodiment editing~\cite{chen2024rovi,chen2024mirage,ji2025oxe}.
We use dataset-specific input and output modules around a shared action expert.
For deployment, we adopt asynchronous chunking inspired by RTC~\cite{black2025real,black2025training}; our evaluation reports the latency budget of the resulting system separately from task success.

The design focus is therefore the residual variable-view video representation and its layer-aligned token interface. Dataset-specific projections and asynchronous execution build on established approaches; our comparisons evaluate their role within this complete policy.

\section{Method}
\label{sec:method}

\subsection{Problem and design}
\label{subsec:method-setup}

\begin{figure}[t]
\centering
\setlength{\parskip}{0pt}
\setlength{\abovecaptionskip}{8pt}
\includegraphics[width=\linewidth]{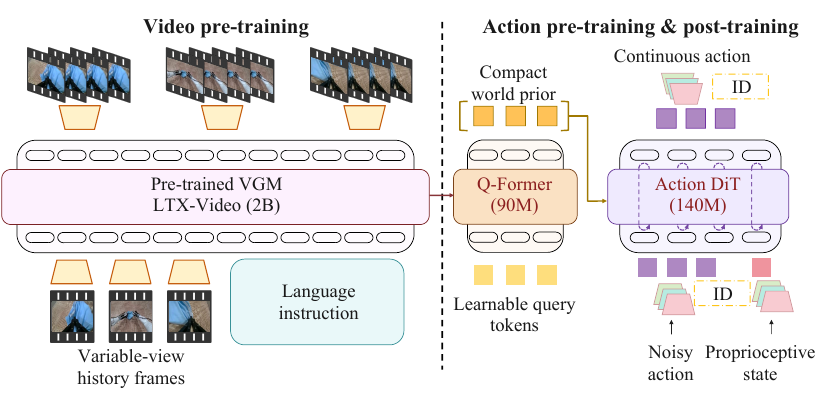}\par\nointerlineskip
\caption{\textbf{Overview of \method{}.} The video backbone is first adapted on robot and human videos with one to three camera views (left). A layer-aligned Q-Former then compresses its features into fixed-length tokens that condition the action DiT, while dataset-specific projections map native states and actions to and from the shared expert (right). Action pretraining learns this interface with the video backbone frozen; target-domain post-training updates all modules using video and action supervision.}\label{fig:pipeline}
\end{figure}

At control step $t$, the policy receives RGB observations $\mathcal{I}_t=\{I_t^{(v)}\}_{v=1}^{V_t}$, an instruction $l$, and proprioceptive state $s_t$.
It predicts an action chunk $\mathbf{a}_{t:t+H-1}\in\mathbb{R}^{H\times d_a}$, with $H=16$ and a dataset-dependent action dimension $d_a$.
Our design separates a variable-view visual encoder from native action decoding through a fixed-length token sequence (Fig.~\ref{fig:pipeline}).
The video prediction horizon used for adaptation is $K=24$ future frames at $256\times256$ resolution; it is distinct from the executable action horizon $H$.

\subsection{Variable-view video representations}
\label{subsec:method-arch}

We initialize the visual backbone from LTX-Video~\cite{hacohen2024ltx}.
Its video VAE compresses inputs by $16\times$ spatially and $8\times$ temporally.
A single observed frame per available camera, together with the instruction, provides the context for learning future-video representations.

\begin{figure}[t]
\centering
\includegraphics[width=\linewidth]{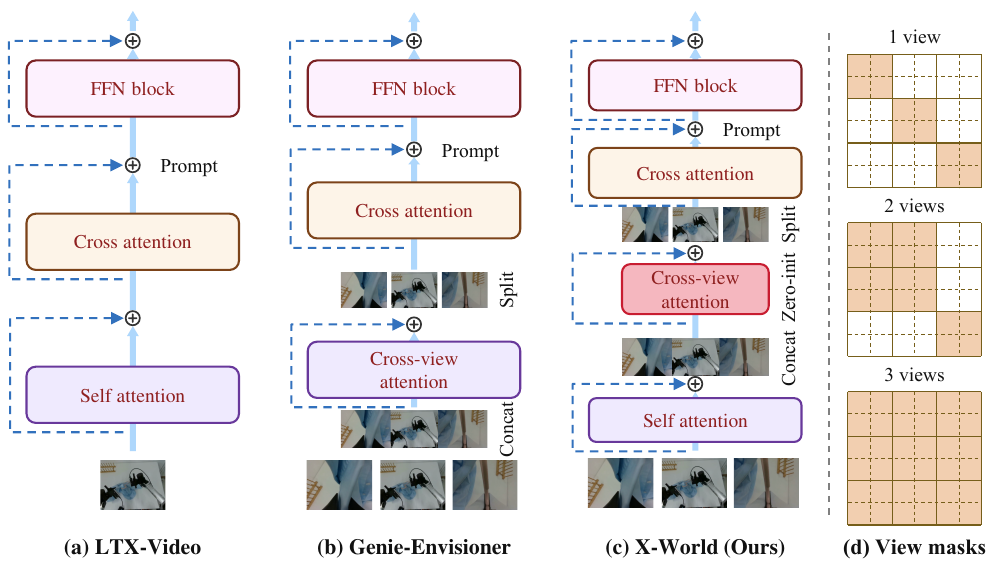}\par
\setlength{\abovecaptionskip}{8pt}
\caption{\textbf{Cross-view adaptation of the video transformer.} (a)~LTX-Video~\cite{hacohen2024ltx} applies self-attention to a single view. (b)~Genie-Envisioner~\cite{liao2025genie} fuses concatenated views through cross-view attention. (c)~\method{} retains pretrained per-view self-attention and learns cross-view fusion through a separate residual branch. Zero initialization preserves the pretrained computation at the start of adaptation while allowing cross-camera interactions to be learned. (d) Training-time masks for one, two, and three available views. Shaded entries allow attention; replicated slots are isolated from available views, supporting joint training across camera counts.}
\label{fig:genie_mask}
\end{figure}

\paragraph{Residual cross-view fusion.}
We retain the pretrained per-view self-attention path and append cross-view attention through an added residual branch (Fig.~\ref{fig:genie_mask}).
Writing $u_\ell$ for the features on that path, the cross-view update has the form
\begin{equation}
\widetilde{u}_\ell=u_\ell+C_\ell(u_\ell;M),\qquad
C_\ell(\cdot;M)=0\ \text{at initialization},
\label{eq:crossview}
\end{equation}
where $M$ masks disallowed interactions between view slots.
The zero initialization makes the added cross-view update vanish initially.
View embeddings encode view identity, and training randomizes the view order.
These operations support heterogeneous camera inputs without replacing the original self-attention with a joint attention operation over every view.

\paragraph{View slots and time coordinates.}
Training uses three view slots.
For a sample with one or two cameras, an existing view is replicated into the remaining slots and independent diffusion noise is applied to the replicas.
The cross-view mask blocks interactions between original and replicated slots in both directions.
The replicas therefore supply additional noisy video inputs without introducing cross-view messages into the original slots.
Per-sample frame rates scale temporal RoPE coordinates~\cite{su2024roformer}; the exact interpolation rule is given in Sec.~\ref{sec:supp-impl-fps}.
The evaluated setting supports one to three views, rather than an unbounded number of cameras.

\paragraph{Layer-aligned token aggregation.}
Let $\{h_\ell\}_{\ell=1}^{L}$ be the exported hidden states of the video transformer's $L$ blocks.
A Q-Former refines learnable queries through a corresponding stack of cross-attention blocks:
\begin{equation}
z_\ell=Q_\ell(z_{\ell-1},h_\ell),\qquad z=z_L.
\label{eq:qformer}
\end{equation}
Here $Q_\ell$ denotes the query-update block attending to video features $h_\ell$.
The final token sequence $z$ has fixed length, independent of the number of video tokens or available views.
This fixes the context length seen by the action expert; the cost of computing the video features and Q-Former can still depend on the input.

\subsection{Shared action expert and training}
\label{subsec:method-training}

The action expert is a DiT-style transformer~\cite{peebles2023scalable}.
It applies self-attention to state and action tokens and cross-attention to $z$.
Following the use of dataset-specific interfaces in X-VLA~\cite{xvla}, each dataset has a state projector, an input projection for noisy actions, and an output head.
The shared transformer operates between these modules, while each output head decodes actions in its dataset's native parameterization.

\begin{table}[t]
\setlength{\abovecaptionskip}{0pt}
\centering\small
\caption{Three-stage training of \method{}. Stage~1 adapts the video backbone and view modules through video prediction on robot and human data. Stage~2 freezes the backbone and trains the Q-Former, action expert, and native interfaces on action-labeled trajectories. Stage~3 jointly optimizes video and action objectives on LIBERO or real-world demonstrations, updating all modules.}
\label{tab:training_phases}
\setlength{\tabcolsep}{3pt}
\begin{tabularx}{\linewidth}{@{}>{\raggedright\arraybackslash}p{.18\linewidth}>{\raggedright\arraybackslash}p{.21\linewidth}>{\raggedright\arraybackslash}p{.21\linewidth}>{\raggedright\arraybackslash}X@{}}
\toprule
Stage & Data & Loss & Updated modules \\
\midrule
1: Video adaptation & Robot/human videos & $\mathcal{L}_{\rm vid}$ & Video backbone and view modules \\
2: Action pretraining & Robot trajectories & $\mathcal{L}_{\rm act}$ & Q-Former, action expert, native projections \\
3: Post-training & Target trajectories & $\mathcal{L}_{\rm vid}+\mathcal{L}_{\rm act}$ & All modules \\
\bottomrule
\end{tabularx}
\end{table}

\paragraph{Stage 1: Video adaptation.}
We adapt the video backbone on robot and human videos.
For future-video latents $\mathbf{y}$, noise $\boldsymbol{\epsilon}\sim\mathcal{N}(0,\mathbf{I})$, and scheduler-sampled noise level $\sigma$, we set $\mathbf{y}_\sigma=(1-\sigma)\mathbf{y}+\sigma\boldsymbol{\epsilon}$ and optimize
\begin{equation}
\mathcal{L}_{\mathrm{vid}}=
\mathbb{E}\!\left[\left\|v_\phi(\mathbf{y}_\sigma,\sigma,\mathcal{I}_t,l)
-(\boldsymbol{\epsilon}-\mathbf{y})\right\|_2^2\right].
\end{equation}
\Needspace{4\baselineskip}
Caption dropout replaces some instructions with an empty-prompt embedding for classifier-free guidance.
The training mixture includes the robot and human video sources listed in Table~\ref{tab:supp_data_stage1}.
The recorded training budget is seven days on 64 H100 GPUs.

\paragraph{Stage 2: Action-only pretraining.}
We freeze the video backbone and update the Q-Former, action expert, and dataset-specific modules with action supervision.
Dataset $d$ is sampled with probability proportional to $N_d^{1/2}$, where $N_d$ is its number of sampleable transitions; all base mixture weights are one.
For a ground-truth action chunk $\mathbf{a}$, noise $\boldsymbol{\epsilon}_a$, and noise level $\sigma_a$, the flow-matching objective is
\begin{equation}
\begin{split}
\mathbf{a}'&=(1-\sigma_a)\mathbf{a}+\sigma_a\boldsymbol{\epsilon}_a,\\
\mathcal{L}_{\mathrm{act}}&=
\mathbb{E}\!\left[\left\|v_\theta(\mathbf{a}',\sigma_a,z,s_t)
-(\boldsymbol{\epsilon}_a-\mathbf{a})\right\|_2^2\right].
\end{split}
\end{equation}
Freezing prevents action gradients from updating the video backbone during this stage.
The data sources are listed in Table~\ref{tab:supp_data_stage2}; the recorded budget is three days on 64 H100 GPUs.

\paragraph{Stage 3: Target-domain post-training.}
We post-train on LIBERO demonstrations for simulation or on collected teleoperation trajectories for the real-world platform.
All model modules are updated with the joint objective
$\mathcal{L}_{\mathrm{post}}=\mathcal{L}_{\mathrm{vid}}+\mathcal{L}_{\mathrm{act}}$.
This stage adapts both the visual representation and action generation to the target domain.

\subsection{Feature extraction and asynchronous execution}
\label{subsec:method-inference}

At control time, we VAE-encode the current views, fix the observed-frame latents, and initialize the unobserved future slots with noise.
One LTX forward pass at the highest-noise endpoint ($\sigma=1$, diffusion timestep 1000) exports features, with observed-frame tokens assigned timestep zero.
The Q-Former and a four-step flow action expert then predict the action chunk.
No iterative video denoising or video decoding lies on this control path.

Asynchronous real-time chunking~\cite{black2025real,black2025training} overlaps a policy query with execution of committed actions.
A FIFO queue $\mathcal{Q}$ stores these actions.
When its remaining length $R=|\mathcal{Q}|$ reaches the threshold $P$, the policy reuses the remaining actions as a committed prefix and predicts a refreshed chunk.
After dropping the overlapping prefix, it appends up to $E$ newly predicted suffix actions.
Here $E$ denotes the appended suffix length, not the prefix length.
For control frequency $f_c$ and refresh latency $T_{\mathrm{refresh}}$, the nominal deadline condition is
\begin{equation}
T_{\mathrm{refresh}} < P/f_c.
\label{eq:deadline}
\end{equation}
With $P=6$ and $f_c=30$\,Hz, the prefix covers 200\,ms.
This budget must cover the actual latency of each query for uninterrupted execution.

Training exposes the action expert to committed prefixes by sampling
$\delta\sim\mathrm{Uniform}(\{0,\ldots,8\})$.
The first $\delta$ action positions remain noise-free and the suffix is noised:
\begin{equation}
\mathbf{a}'_i=(1-\sigma_{a,i})\mathbf{a}_i+\sigma_{a,i}\boldsymbol{\epsilon}_{a,i},
\qquad
\sigma_{a,i}=
\begin{cases}
0,&i\leq\delta,\\
\sigma_a,&i>\delta.
\end{cases}
\end{equation}
Thus the action expert receives a known prefix during both training and deployment.

\section{Experiments}
\label{sec:exp}

We evaluate task performance on LIBERO and LIBERO-Plus, then present the real-world platform, tasks, and rollouts before analyzing the training and deployment choices.
Future-video prediction examples are provided in Sec.~\ref{sec:supp-additional-video-generation}.

\subsection{Simulation results}
\label{subsec:exp-sim}

\paragraph{LIBERO.}
We post-train one unified policy on demonstrations from all four LIBERO suites~\cite{liu2023libero}.
Inputs are the language instruction, synchronized third-person and wrist RGB images, and proprioceptive state.
Evaluation uses 50 episodes for each of 10 tasks per suite, giving 500 rollouts per suite.
Table~\ref{tab:main_results} reports the complete reference comparison~\cite{Diffusionpolicy,kim2024openvla,qu2025spatialvla,zhao2025cotvla,pertsch2025fast,bjorck2025gr00t,black2024pi0,lv2025f1,internvlam1,liang2025discrete,liao2025genie,pi05,OFT,xvla}.
These reference results use different pretraining data and compute budgets.

\begin{table}[t]
\setlength{\abovecaptionskip}{0pt}
\centering
\small
\renewcommand{\arraystretch}{1.32}
\caption{LIBERO success rates (\%) on the Spatial, Object, Goal, and Long suites. Average is the mean of the four suite scores. Best results are bold with pink backgrounds; second-best results are underlined with lavender backgrounds.}
\label{tab:main_results}\label{tab:supp_main_results}
\setlength{\tabcolsep}{2pt}
\begin{tabular*}{\linewidth}{@{\extracolsep{\fill}}>{\raggedright\arraybackslash}p{.49\linewidth}cccc|c@{}}
\toprule
Method & \hphantom{$\uparrow$}Spatial$\uparrow$ & \hphantom{$\uparrow$}Object$\uparrow$ & \hphantom{$\uparrow$}Goal$\uparrow$ & \hphantom{$\uparrow$}Long$\uparrow$ & Average \\
\midrule
Diffusion Policy~\cite{Diffusionpolicy}
& 78.5 & 87.5 & 73.5 & 64.8 & 76.1 \\
OpenVLA~\cite{kim2024openvla}
& 84.7 & 88.4 & 79.2 & 53.7 & 76.5 \\
SpatialVLA~\cite{qu2025spatialvla}
& 88.2 & 89.9 & 78.6 & 55.5 & 78.1 \\
CoT-VLA~\cite{zhao2025cotvla}
& 87.5 & 91.6 & 87.6 & 69.0 & 83.9 \\
$\pi_0$-Fast~\cite{pertsch2025fast}
& 96.4 & 96.8 & 88.6 & 60.2 & 85.5 \\
GR00T-N1~\cite{bjorck2025gr00t}
& 94.4 & 97.6 & 93.0 & 90.6 & 93.9 \\
$\pi_0$~\cite{black2024pi0}
& 98.0 & 96.8 & 94.4 & 88.4 & 94.4 \\
F1~\cite{lv2025f1}
& 98.2 & 97.8 & 95.4 & 91.3 & 95.7 \\
InternVLA-M1~\cite{internvlam1}
& 98.0 & \second{99.0} & 93.8 & 92.6 & 95.9 \\
Discrete Diffusion VLA~\cite{liang2025discrete}
& 97.2 & 98.6 & 97.4 & 92.0 & 96.3 \\
Genie-Envisioner~\cite{liao2025genie}
& 98.2 & 97.6 & 95.8 & 94.4 & 96.5 \\
$\pi_{0.5}$~\cite{pi05}
& \second{98.8} & 98.2 & \second{98.0} & 92.4 & 96.9 \\
GR00T-N1.6~\cite{gear2025gr00tn16}
& 97.7 & 98.5 & 97.5 & 94.4 & 97.0 \\
OpenVLA-OFT~\cite{OFT}
& 97.6 & 98.4 & 97.9 & 94.5 & 97.1 \\
X-VLA~\cite{xvla}
& 98.2 & 98.6 & 97.8 & \first{97.6} & \second{98.1} \\
\midrule
\method{} & \first{99.6} & \first{99.4} & \first{99.0} & \second{95.8} & \first{98.5} \\
\bottomrule
\end{tabular*}
\end{table}

\method{} reaches 98.5\% mean success.
Its largest numerical advantage over X-VLA is on Spatial (99.6\% versus 98.2\%), whereas X-VLA performs better on Long (97.6\% versus 95.8\%).
The 0.4-point average gap is small at this saturation level, motivating the distribution-shift evaluation and within-framework ablations below.

\paragraph{Zero-shot LIBERO-Plus.}
We evaluate the LIBERO-post-trained checkpoint on LIBERO-Plus~\cite{fei2025liberoplus} without further adaptation.
Table~\ref{tab:libero_plus} gives the full seven-shift comparison~\cite{kim2024openvla,black2024pi0,OFT,hung2025nora,cen2025worldvla,bu2025univla,pertsch2025fast,tan2025interactive}.
The reported total is 76.5\%, compared with 69.6\% for OpenVLA-OFT.
The largest category difference is under robot perturbations (72.5\% versus 31.9\%); language and layout also improve by 14.8 and 10.5 points.
Camera, background, and noise success are lower by 3.8, 7.5, and 16.1 points, respectively.
The aggregate gain therefore reflects selective robustness: accommodating different camera counts does not by itself confer invariance to changed camera poses or corrupted images.

\begin{table}[t]
\setlength{\abovecaptionskip}{0pt}
\centering
\small
\renewcommand{\arraystretch}{1.12}
\caption{Zero-shot success rates (\%) across seven LIBERO-Plus distribution shifts~\cite{fei2025liberoplus}. Cam., Lang., and Bg. denote camera, language, and background changes, respectively. The w and m variants use no wrist input and mixed-data supervised fine-tuning, respectively. OpenVLA-OFT+ is excluded because it is trained on LIBERO-Plus. Best results are bold with pink backgrounds; second-best results are underlined with lavender backgrounds.}
\label{tab:libero_plus}\label{tab:supp_libero_plus}
\setlength{\tabcolsep}{2pt}
\begin{tabular*}{\linewidth}{@{\extracolsep{\fill}}>{\raggedright\arraybackslash}p{.38\linewidth}ccccccc|c@{}}
\toprule
Method & Cam. & Robot & Lang. & Light & Bg. & Noise & Layout & Total \\
\midrule
OpenVLA~\cite{kim2024openvla}
& 0.8 & 3.5 & 23.0 & 8.1 & 34.8 & 15.2 & 28.5 & 15.6 \\
$\pi_0$~\cite{black2024pi0}
& 13.8 & 6.0 & 58.8 & 85.0 & 81.4 & \first{79.0} & 68.9 & 53.6 \\
OpenVLA-OFT~\cite{OFT}
& \second{56.4} & 31.9 & 79.5 & 88.7 & \second{93.3} & 75.8 & \second{74.2} & \second{69.6} \\
OpenVLA-OFT$_{\mathrm{w}}$~\cite{OFT}
& 10.4 & 38.7 & 70.5 & 76.8 & \first{93.6} & 49.9 & 69.9 & 55.8 \\
NORA~\cite{hung2025nora}
& 2.2 & 37.0 & 65.1 & 45.7 & 58.6 & 12.8 & 62.1 & 39.0 \\
WorldVLA~\cite{cen2025worldvla}
& 0.1 & 27.9 & 41.6 & 43.7 & 17.1 & 10.9 & 38.0 & 25.0 \\
UniVLA~\cite{bu2025univla}
& 1.8 & \second{46.2} & 69.6 & 69.0 & 81.0 & 21.2 & 31.9 & 43.9 \\
$\pi_0$-Fast~\cite{pertsch2025fast}
& \first{65.1} & 21.6 & 61.0 & 73.2 & 73.2 & 74.4 & 68.8 & 61.6 \\
RIPT-VLA~\cite{tan2025interactive}
& 55.2 & 31.2 & 77.6 & 88.4 & 91.6 & 73.5 & \second{74.2} & 68.4 \\
OpenVLA-OFT$_{\mathrm{m}}$~\cite{OFT}
& 55.6 & 21.7 & \second{81.0} & \second{92.7} & 91.0 & \second{78.6} & 68.7 & 67.9 \\
\midrule
\method{}
& 52.6 & \first{72.5} & \first{94.3} & \first{95.4} & 85.8 & 59.7 & \first{84.7} & \first{76.5} \\
\bottomrule
\end{tabular*}
\end{table}

\Needspace{16\baselineskip}
\subsection{Real-world setup and tasks}
\label{subsec:exp-real}
\label{sec:supp-realworld}

\begingroup
\setlength{\intextsep}{4pt}
\setlength{\columnsep}{12pt}
\makeatletter\@nobreakfalse\makeatother
\everypar{}
\begin{wrapfigure}{r}{0.49\textwidth}
\centering
\setlength{\abovecaptionskip}{1pt}
\includegraphics[width=\linewidth]{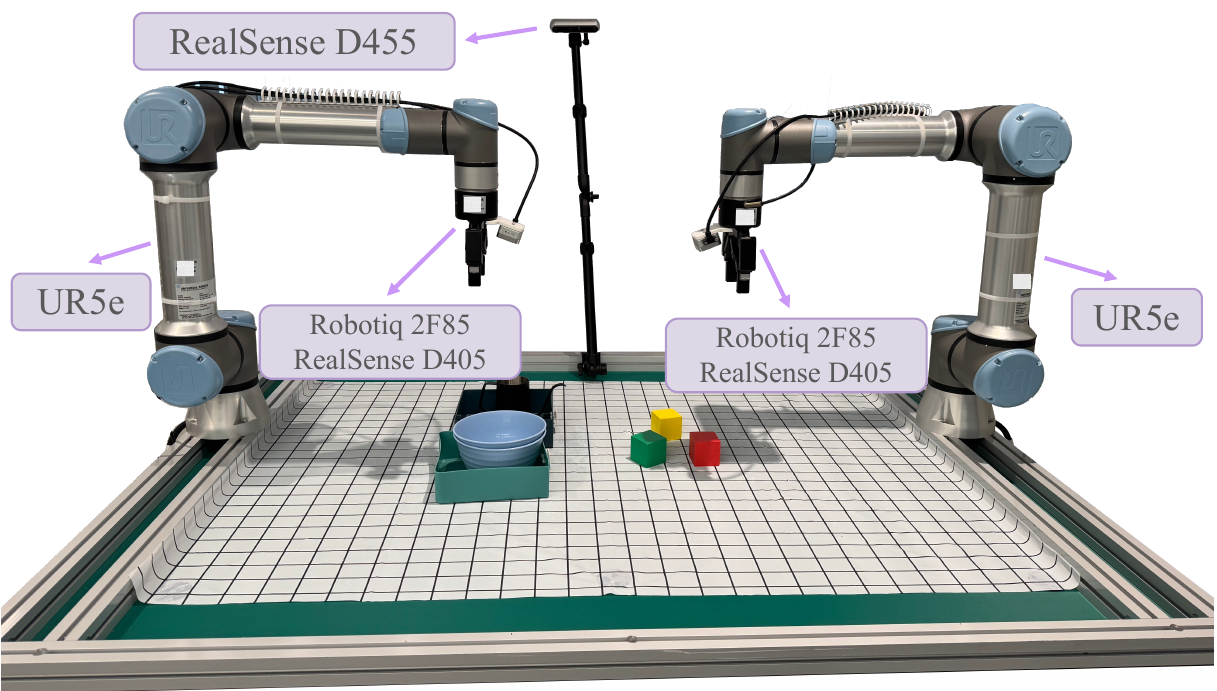}
\caption{\textbf{Dual-arm evaluation platform.} Two UR5e arms use Robotiq 2F85 grippers and wrist-mounted RealSense D405 cameras; a shared D455 camera supplies the third-person view.}
\label{fig:real_world_setup}
\vspace{7pt}
\end{wrapfigure}

\noindent\textbf{Platform and scene.}
We evaluate \method{} on the dual-arm platform in Fig.~\ref{fig:real_world_setup}.
Two wrist cameras and a shared third-person camera provide synchronized RGB observations of the tabletop workspace.

\noindent\textbf{Data and evaluation protocol.}
Each task provides 30 minutes of VR teleoperation demonstrations with synchronized three-view observations.
Data recording and action execution run at 30\,Hz.
Actions are relative joint position targets, and absolute joint positions form the proprioceptive input.
Each policy is evaluated on 20 trials per task using the same 20 predefined initial placements, camera views, control frequency, and task definitions.
\par\WFclear
\endgroup

\noindent\begin{minipage}{\linewidth}
The six evaluated tasks require sequential or simultaneous coordination:
\begingroup
\setlength{\leftmargini}{1.2em}
\setlength{\labelsep}{0.5em}
\begin{itemize}
    \item \textbf{Stack bowl and block.} Stack a side bowl onto a center bowl, then place a block into it.
    \item \textbf{Put bottle into trash can.} Pick up scattered bottles and place them in the trash can.
    \item \textbf{Pour water from teapot.} Pour water into a cup, then place the teapot and cup on a tray.
    \item \textbf{Move bowl into a tray.} Move a bowl to the center with one arm, then transfer it into a tray with the other.
    \item \textbf{Stack two blocks.} Move blocks from the table to a tray and stack them.
    \item \textbf{Clear the table.} Place all objects into a drawer.
\end{itemize}
\endgroup
\end{minipage}

\begin{figure}[t]
  \centering
  \setlength{\parskip}{0pt}
  \setlength{\abovecaptionskip}{8pt}
  \includegraphics[width=\linewidth]{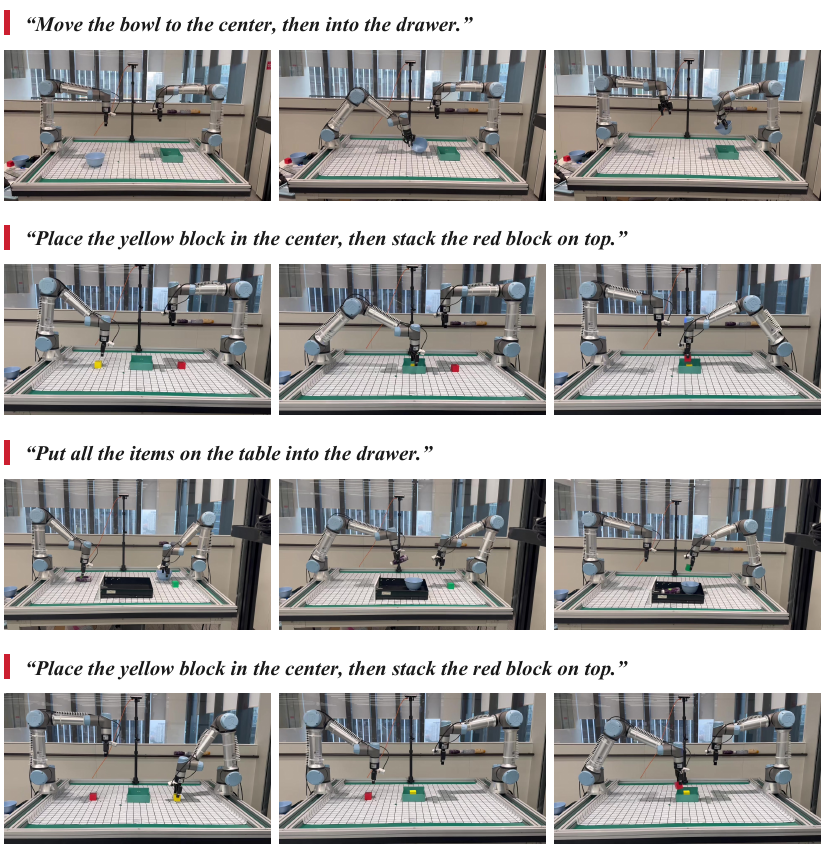}\par\nointerlineskip
  \caption{\textbf{Real-world manipulation examples.} Each row corresponds to a separate rollout of the language instruction above it, with three third-person snapshots arranged in execution order from left to right.}\label{fig:realworld_vis}
\end{figure}

\paragraph{Qualitative execution.}
Figure~\ref{fig:realworld_vis} presents representative real-world rollouts, with three ordered snapshots per row.
The instruction above each row identifies that example.
These selected rollouts illustrate the workspace and manipulation sequence; the success rates below are computed from the evaluation trials.
The supplementary package provides six task videos (Sec.~\ref{sec:supp-videos}).

\subsection{Real-world results}
\label{subsec:exp-real-results}

\begin{figure}[t]
  \centering
  \setlength{\parskip}{0pt}
  \setlength{\abovecaptionskip}{8pt}
  \includegraphics[width=\linewidth]{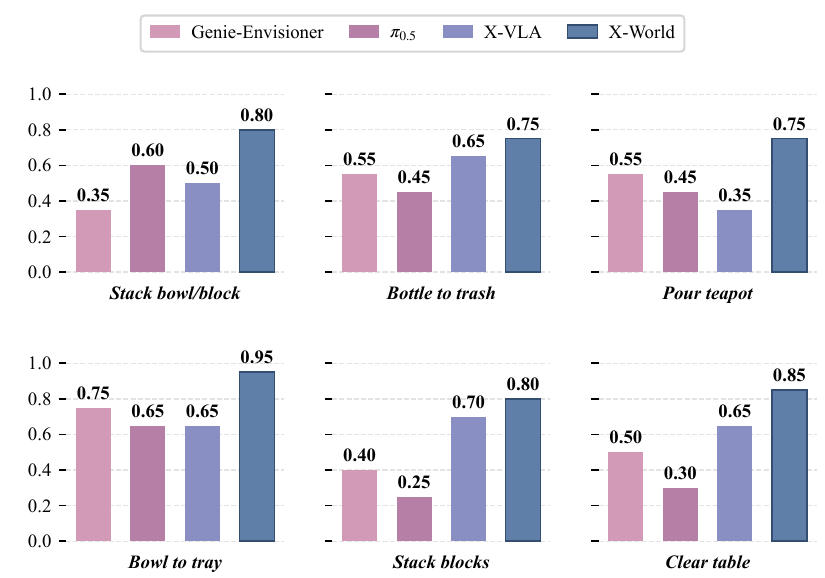}\par\nointerlineskip
  \caption{\textbf{Real-world task success.} Bars show success fractions over 20 trials per policy and task on six dual-arm UR5e tasks under a shared evaluation protocol and identical initial placements. \method{} achieves the highest observed success on every task.}\label{fig:realworld_results}
\end{figure}

Figure~\ref{fig:realworld_results} reports the per-task success rates for \method{} and the three baseline policies~\cite{liao2025genie,pi05,xvla}.
\method{} achieves the highest point estimate on each task, with success rates between 75\% and 95\%.
Across six tasks it succeeds in 98 of 120 trials, giving an 81.7\% task mean.
X-VLA achieves 70/120 (58.3\%), Genie-Envisioner 62/120 (51.7\%), and $\pi_{0.5}$ 54/120 (45.0\%).
The comparison is based on 20 trials per task; the available aggregate counts do not determine training-seed variability or paired trial outcomes.

\subsection{Ablation studies}
\label{subsec:exp-ablation}

\paragraph{Training stages.}
We compare the four combinations of video adaptation and action pretraining, with target-domain post-training and the LIBERO evaluation protocol held fixed.
With video adaptation present, action pretraining increases mean success from 97.4\% to 98.5\%, and Long success from 94.2\% to 95.8\%.
Without video adaptation, action pretraining instead lowers mean success from 92.1\% to 89.5\%, and Long success from 81.2\% to 72.4\%.
The two stages therefore do not behave as independent, additive improvements in these results.
The comparison is consistent with an interaction between visual adaptation and action pretraining, but does not identify the cause of the reversal.

\begin{table}[t]
\setlength{\abovecaptionskip}{0pt}
\centering
\small
\renewcommand{\arraystretch}{1.12}
\caption{LIBERO success rates (\%) for ablations of training stages, model components, and video loss. Stage~1 is video adaptation and Stage~2 is action pretraining; Average is the mean across the four suites. Best results are bold with pink backgrounds; second-best results are underlined with lavender backgrounds.}
\label{tab:ablation}
\setlength{\tabcolsep}{4.2pt}
\renewcommand{\arraystretch}{1.05}
\begin{tabular*}{\linewidth}{@{\extracolsep{\fill}}lcccc|c@{}}
\toprule
Setting & \hphantom{$\uparrow$}Spatial$\uparrow$ & \hphantom{$\uparrow$}Object$\uparrow$ & \hphantom{$\uparrow$}Goal$\uparrow$ & \hphantom{$\uparrow$}Long$\uparrow$ & Average \\
\midrule
w/o Stage~1 & 95.4 & 97.6 & 92.4 & 72.4 & 89.5 \\
w/o Stage~2 & 99.2 & 98.8 & 97.2 & 94.2 & 97.4 \\
w/o Stage~1 and Stage~2 & 97.4 & 96.2 & 93.4 & 81.2 & 92.1 \\
Shared projections & 98.6 & 98.4 & 98.2 & 94.4 & 97.4 \\
w/o Q-Former & \first{99.8} & 98.6 & \second{98.6} & 94.6 & 97.9 \\
w/o video loss & \first{99.8} & \first{100.0} & 97.2 & \first{96.2} & \second{98.3} \\
\midrule
Full model & \second{99.6} & \second{99.4} & \first{99.0} & \second{95.8} & \first{98.5} \\
\bottomrule
\end{tabular*}
\end{table}

\paragraph{Q-Former and native action projections.}
Replacing the Q-Former with direct access to last-layer video features gives 97.9\% mean success, compared with 98.5\% for the full model.
This comparison changes both feature aggregation and compression.
The Q-Former takes 5\,ms in the full-model profile; a matched latency measurement for the direct-feature variant is unavailable.
Shared state and action projections yield 97.4\% mean success, 1.1 points below the dataset-specific design.

\paragraph{Video loss during post-training.}
Removing the video loss gives 98.3\% mean success, including a higher Long score (96.2\% versus 95.8\%).
This small average difference provides limited evidence for a control benefit from the joint video objective.

\paragraph{Cross-view components.}
Removing residual cross-view attention lowers LIBERO success from 98.5\% to 83.1\%; removing view embeddings lowers it to 93.3\%.
We additionally evaluate \textit{Fold Towel}, which requires flattening a cluttered towel and then folding it.
The full model succeeds in 12/20 trials, compared with 0/20 without residual cross-view attention and 2/20 without view embeddings.
These component removals are distinct from the test-time view-availability analysis below.

\subsection{View availability and real-time execution}
\label{subsec:exp-runtime}
\label{sec:supp-additional-ablation}

\begin{table}[t]
\centering
\small
\renewcommand{\arraystretch}{1.12}
\caption{Single-H100 inference latency (ms). Component timings exclude 14 ms of model-side overhead included in the 46 ms cached response; T5 encoding adds 13 ms on the first call. The 91 ms deployed refresh additionally includes network transfer, data processing, and control-side overhead.}
\label{tab:speed_breakdown}\label{tab:runtime}\label{tab:supp_runtime}
\setlength{\tabcolsep}{2pt}
\begin{tabular*}{\linewidth}{@{\extracolsep{\fill}}lr@{\hspace{18pt}}lr@{}}
\toprule
\multicolumn{2}{c}{Model timing} & \multicolumn{2}{c}{Deployment timing} \\
\midrule
T5 encoder (first call) & 13 & Cached model response & 46 \\
VAE & 7 & Network transfer & 32 \\
Video diffusion & 12 & Data processing & 7 \\
Q-Former & 5 & Control-side overhead & 6 \\
Action expert & 8 & End-to-end refresh & 91 \\
\bottomrule
\end{tabular*}%
\end{table}

\begin{figure}[t]
  \centering
  \setlength{\parskip}{0pt}
  \setlength{\abovecaptionskip}{8pt}
  \includegraphics[width=\linewidth]{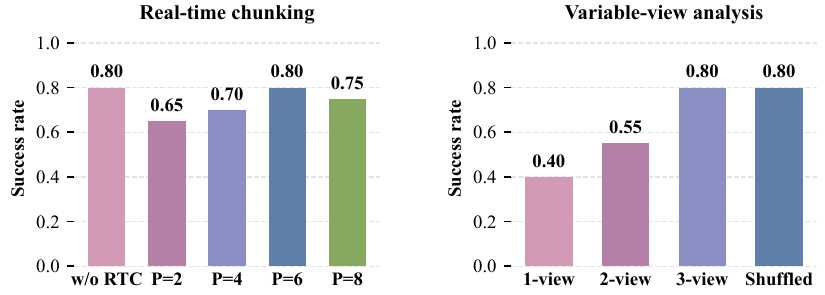}\par\nointerlineskip
  \caption{\textbf{Action-prefix and view-availability analyses.} The left panel compares inference without real-time chunking (w/o RTC) with prefix lengths~\mbox{$P=2,4,6,8$}. The right panel uses the same checkpoint to compare one, two, and three camera views and a shuffled three-view order. Bars report success fractions.}\label{fig:supp_additional_results}
\end{figure}

\paragraph{Available views.}
We use the same model checkpoint for all view-availability settings.
One-view evaluation retains the third-person camera; two-view evaluation removes one wrist camera; the shuffled setting permutes the three-camera input order.
The reported success rates are 40\%, 55\%, 80\%, and 80\% for one, two, three, and shuffled views, respectively (Fig.~\ref{fig:supp_additional_results}, right).
Additional views improve the point estimate, while the tested order permutation leaves it unchanged.
The aggregate sweep records do not specify task selection, trial denominators, or seed variation, so these results are descriptive.

\paragraph{Action-prefix sweep.}
Success is 80\% without RTC and 65\%, 70\%, 80\%, and 75\% for $P=2,4,6,8$ (Fig.~\ref{fig:supp_additional_results}, left).
The reported point estimates favor $P=6$ among the tested prefixes.
Success alone does not measure execution continuity, as the no-RTC setting also reaches 80\%.

\paragraph{Runtime breakdown.}
Table~\ref{tab:speed_breakdown} reports the module and deployment timing breakdown for the single-H100 system with $H=16$.
The VAE, video backbone, Q-Former, and action expert sum to 32\,ms.
Including 14\,ms of model-side overhead gives a 46\,ms cached response; the first call adds 13\,ms of T5 encoding.
Network transfer, data processing, and control-side overhead raise the deployed refresh to 91\,ms.
These timings differ from the 30\,Hz clock that executes queued actions.

At 30\,Hz, prefixes of two, four, six, and eight actions cover approximately 67, 133, 200, and 267\,ms.
Only the first is below the reported 91\,ms refresh.
With $P=6$, the committed prefix provides a nominal margin of 109\,ms.
This budget explains the deployment choice without establishing a bound on every query or an observed queue-underflow rate.

\section{Discussion and conclusion}
\label{sec:conclusion}

\method{} transfers video representations to robot control through residual multi-view adaptation and a fixed-length token interface.
Removing video adaptation or cross-view fusion produces the largest observed success drops among the tested removals.
The training-stage comparison also exposes an interaction: action pretraining helps after video adaptation, but reduces performance when adaptation is skipped.
The distribution-shift results are similarly selective, with gains on robot and language perturbations accompanied by losses on camera and noise shifts.
Together, these findings motivate treating the visual and action interfaces as separate design choices when transferring a video prior.

\paragraph{Limitations.}
Evaluation covers LIBERO, its distribution-shift extension, and one real-world robot platform; it does not establish transfer to an unseen embodiment or a scaling law.
The training-stage ablations change the available supervision as well as the training procedure, so they do not isolate the benefit of freezing the backbone.
The Q-Former comparison uses last-layer features, confounding compression with access to multiple layers; its 0.6-point average difference requires repeated runs and a matched feature baseline before drawing a stronger conclusion.
Baseline pretraining data and compute are not controlled, and the available results do not quantify training-seed variability.
The LIBERO-Plus category denominators and aggregate weighting are unspecified, limiting audit of the reported total score.
Real-world success estimates are based on 20 trials per task.
The latency profile provides a nominal queue budget; tail latency, queue underflow, and action-boundary continuity remain unquantified.
\label{sec:main_end}

\subsection*{Ethics statement}

This work uses public human and robot video datasets and self-collected robot
teleoperation data for research on language-instructed manipulation. Public
data are used according to their licenses. Teleoperation recordings are
handled as potentially identifying data, and any redistribution should
preserve the applicable consent and privacy requirements. Physical robot
experiments require supervised operation, workspace safeguards, and an
emergency stop; success in simulation or a qualitative rollout does not
establish general hardware safety.

\subsection*{Reproducibility statement}

The paper reports the training objectives, updated modules, input resolution,
prediction horizons, feature-extraction path, and evaluation settings.
The Appendix lists data sources, frame-rate interpolation, native action
representations, and the available aggregate results.
Figure scripts use those aggregates and the original qualitative frames.
Full model dimensions, optimization configurations, post-training budgets,
and per-run evaluation logs are not supplied here; these omissions limit
independent reproduction and uncertainty analysis.


\setlength{\bibsep}{7.5pt plus 1pt minus 0.5pt}

\appendix

\makeatletter
\renewcommand{\section}{\@startsection{section}{1}{\z@}{-1.5ex plus
    -0.4ex minus -.2ex}{1.0ex plus 0.2ex minus 0.2ex}{\large\sc\raggedright}}
\renewcommand{\subsection}{\@startsection{subsection}{2}{\z@}{-1.3ex plus
    -0.4ex minus -.2ex}{0.6ex plus 0.15ex minus 0.15ex}{\normalsize\sc\raggedright}}
\renewcommand{\subsubsection}{\@startsection{subsubsection}{3}{\z@}{-1.1ex plus
    -0.3ex minus -.2ex}{0.4ex plus 0.1ex minus 0.1ex}{\normalsize\sc\raggedright}}
\makeatother

\section{Data and Training Details}
\label{sec:supp-data}

This section summarizes the data sources used across the three training stages of \method{} and the sampling strategy used for heterogeneous action pretraining.

\subsection{Stage 1: Video pre-training}
\label{sec:supp-data-stage1}

We use a mixture of robot videos and human videos to provide diverse visual interaction data for video pre-training. All sources are converted into a unified variable-view format, with up to three views retained for each sample. Table~\ref{tab:supp_data_stage1} summarizes the resulting data sources.

\begin{table}[ht]
\centering
\small
\caption{Datasets for Stage~1 video adaptation. Robot and human videos supply future-frame supervision, while Views gives the number of available camera streams per training example.}
\label{tab:supp_data_stage1}
\setlength{\tabcolsep}{2.2pt}
\renewcommand{\arraystretch}{1.01}
\begin{tabularx}{\linewidth}{@{}>{\raggedright\arraybackslash}p{0.28\linewidth}c>{\raggedright\arraybackslash}X@{}}
\toprule
Dataset & Views & Summary \\
\midrule
RoboTwin 2.0~\cite{chen2025robotwin} & 3 & Domain-randomized dual-arm benchmark designed for robust bimanual manipulation and transfer. \\
CALVIN~\cite{mees2022calvin} & 2 & Long-horizon language benchmark built around static and gripper observations for sequential manipulation. \\
AgiBotWorld~\cite{bu2025agibot} & 3 & Quality-controlled manipulation data collected in realistic settings with broad skill coverage. \\
Behavior-1K~\cite{li2024behavior1k} & 3 & Household simulation benchmark emphasizing realistic environments and long-horizon daily activities. \\
RoboCOIN~\cite{wu2025robocoin} & 3 & Bimanual multi-embodiment data with hierarchical annotations for coordinated manipulation skills. \\
Galaxea~\cite{jiang2025galaxea} & 3 & Open-world robot data from everyday scenes with fine-grained subtask language annotations. \\
Open-X-Embodiment~\cite{o2024open} & 1-3 & Aggregated real-robot data spanning diverse embodiments, camera layouts, and manipulation behaviors. \\
RoboMIND~\cite{wu2024robomind} & 1-3 & Unified multi-embodiment teleoperation data covering diverse tasks, objects, and robot platforms. \\
Ego4D~\cite{grauman2022ego4d} & 1 & Large-scale egocentric daily-life videos capturing diverse activities and environments from a first-person view. \\
Humanoid-Everyday~\cite{zhao2025humanoid} & 1 & Open-world humanoid manipulation data with multimodal sensing for whole-body interaction tasks. \\
\bottomrule
\end{tabularx}
\end{table}

\subsection{Stage 2: Action pre-training}
\label{sec:supp-data-stage2}

We use heterogeneous VLA trajectories collected from multiple benchmarks and robot platforms for action pre-training. These datasets differ in embodiment, observation setup, and heterogeneous state and action interfaces, providing broad supervision across diverse manipulation settings. Table~\ref{tab:supp_data_stage2} summarizes the main VLA data sources.

\begin{table}[ht]
\centering
\small
\caption{Datasets for Stage~2 action pretraining. Robot trajectories provide action supervision, while Views gives the available camera count per example; dataset-specific projections accommodate native state and action formats.}
\label{tab:supp_data_stage2}
\setlength{\tabcolsep}{2.2pt}
\renewcommand{\arraystretch}{1.01}
\begin{tabularx}{\linewidth}{@{}>{\raggedright\arraybackslash}p{0.28\linewidth}c>{\raggedright\arraybackslash}X@{}}
\toprule
Dataset & Views & Summary \\
\midrule
RoboTwin 2.0~\cite{chen2025robotwin} & 3 & Domain-randomized dual-arm trajectories for robust bimanual manipulation and transfer. \\
CALVIN~\cite{mees2022calvin} & 2 & Long-horizon language-conditioned trajectories with static and gripper observations. \\
AgiBotWorld~\cite{bu2025agibot} & 3 & Large-scale manipulation trajectories collected in realistic settings with diverse skills. \\
Behavior-1K~\cite{li2024behavior1k} & 3 & Household embodied benchmark providing long-horizon trajectories in realistic simulation. \\
RoboCOIN~\cite{wu2025robocoin} & 3 & Bimanual multi-embodiment trajectories with hierarchical annotations for coordinated skills. \\
Galaxea~\cite{jiang2025galaxea} & 3 & Open-world manipulation trajectories with fine-grained subtask language supervision. \\
Open-X-Embodiment~\cite{o2024open} & 1-3 & Aggregated cross-embodiment trajectories covering diverse robots, scenes, and tasks. \\
RoboMIND~\cite{wu2024robomind} & 1-3 & Unified multi-embodiment teleoperation trajectories across diverse tasks and platforms. \\
\bottomrule
\end{tabularx}
\end{table}

\subsection{Temperature-scaled multi-dataset sampling}
\label{sec:supp-data-mixing}

To avoid over-sampling the largest sources while still accounting for dataset scale, we use temperature-scaled sampling when drawing training examples from the Stage~2 mixture. The sampling weight of dataset $d$ is defined as
\begin{equation}
w_d\propto \alpha_d\cdot N_d^{1/T},
\end{equation}
where $N_d$ denotes the total number of dataset steps, i.e., the number of sampleable state and action transitions in dataset $d$, $\alpha_d$ is the base mixture weight, and $T$ is the sampling temperature. We set $\alpha_d=1.0$ and use $T=2.0$ for all datasets, so the effective reweighting scales with the square root of the dataset step count before normalization.

\subsection{Stage 3: Real-world post-training data}
\label{sec:supp-data-stage3}

We use self-collected teleoperation data for post-training on the six real-world tasks described in \mainsectionref{sec:supp-realworld}. We collect 30 minutes of demonstrations per task with synchronized three-view observations. The platform, control interface, evaluation protocol, and task definitions are described in \mainsectionref{sec:supp-realworld}.

\section{Implementation Details}
\label{sec:supp-impl}

This section provides additional implementation details for variable-view processing, the token sequence interface, heterogeneous state and action alignment, and asynchronous real-time chunking.

\subsection{Variable-view video processing}
\label{sec:supp-impl-variable-view}

We add a view embedding to encode view identity without assuming a fixed semantic ordering, and randomize view order during training to improve robustness to heterogeneous camera setups. Each training sample is represented with three view slots. When only one or two views are available, we replicate an existing view to fill the missing slots and apply independent diffusion noise to each replica, which avoids padding and prevents wasting computation on masked placeholders. Cross-view attention masks block interactions between replicated and original view slots, ensuring that replicated views do not affect the original views and vice versa.

\subsection{FPS-aware temporal RoPE scaling}
\label{sec:supp-impl-fps}

Our video diffusion backbone builds on LTX-Video~\cite{hacohen2024ltx} and ingests videos collected at different frame rates. To align temporal positional encoding with the physical sampling rate of each sample, we pass the per-sample frame rate $\mathrm{fps}$ to the backbone and use it to construct the RoPE interpolation tuple~\cite{su2024roformer} in both training and inference:
\begin{equation}
\mathrm{rope\_interpolation\_scale}=\Big(\tfrac{r_t}{\mathrm{fps}},\ r_s,\ r_s\Big),
\end{equation}
where $r_t$ and $r_s$ denote the temporal and spatial compression ratios of the LTX-Video VAE. For multi-view inputs, the same sample-level $\mathrm{fps}$ is repeated across views. In the transformer, this tuple scales the temporal, height, and width coordinates before rotary embedding is applied, so clips with different frame rates share the same architecture while their temporal coordinates remain aligned with the actual sampling rate.

\subsection{Q-Former and token sequence interface}
\label{sec:supp-impl-qformer}

The video diffusion backbone produces spatiotemporal features throughout its denoising transformer blocks. We denote the exported hidden states by $\{h_{\ell}\}_{\ell=1}^{L}$ and distill them into a fixed-length token sequence using a layer-aligned Q-Former. At layer $\ell$, the current query tokens cross-attend to the corresponding backbone hidden states $h_{\ell}$ before being passed to the next layer. The final token sequence forms a compact world prior whose size is independent of the number of input views and video tokens, and the action expert attends to this fixed interface when generating action chunks.

\subsection{Action representation and dataset-specific modules}
\label{sec:supp-impl-action}

VLA datasets have heterogeneous state and action interfaces, including different action parameterizations, units, and coordinate frames. Rather than forcing a single shared action space, we use lightweight dataset-specific modules around a shared action-generation backbone, following the design in X-VLA~\cite{xvla}. Each dataset has a state projector, an action input projection for noised action chunks, and an output head that decodes shared representations into executable actions in its native parameterization. For the real-world platform, actions are represented as relative joint position targets, while absolute joint positions are provided as proprioceptive state.

\subsection{Asynchronous real-time chunking}
\label{sec:supp-impl-chunking}

We maintain a FIFO action queue $\mathcal{Q}$ and execute and remove its front action at each control step. Let $R\triangleq |\mathcal{Q}|$ be the remaining queue length, $P$ the prefix threshold, and $E$ the number of newly predicted suffix actions appended after each refresh. When $R\leq P$, the policy uses the current observations and reuses the remaining $R$ actions as a committed prefix to generate a refreshed action chunk. It discards the overlapping $R$ predicted actions and appends the next $E$ suffix actions to $\mathcal{Q}$. The robot continues executing committed actions while the policy updates future actions in the background, decoupling inference updates from action execution and avoiding waiting gaps.

In the reference deployment implementation, $P$ corresponds to prefix\_steps, $E$ corresponds to execution\_steps, and delay is computed as $\min(\text{remain\_steps},\text{prefix\_steps})$. Thus, delay gives the number of remaining queued actions reused as the action prefix and dropped from the overlapping part of the returned prediction.

\section{Additional Video Predictions}
\label{sec:supp-additional}
\label{sec:supp-additional-video-generation}

The complete benchmark comparisons, real-world setup, task definitions, action-prefix and view-availability analyses, and runtime breakdown are included in \mainsectionref{sec:exp}.
This appendix provides the additional qualitative video predictions.

\begin{figure}[t]
  \centering
  \setlength{\parskip}{0pt}
  \setlength{\abovecaptionskip}{8pt}
  \includegraphics[width=\linewidth]{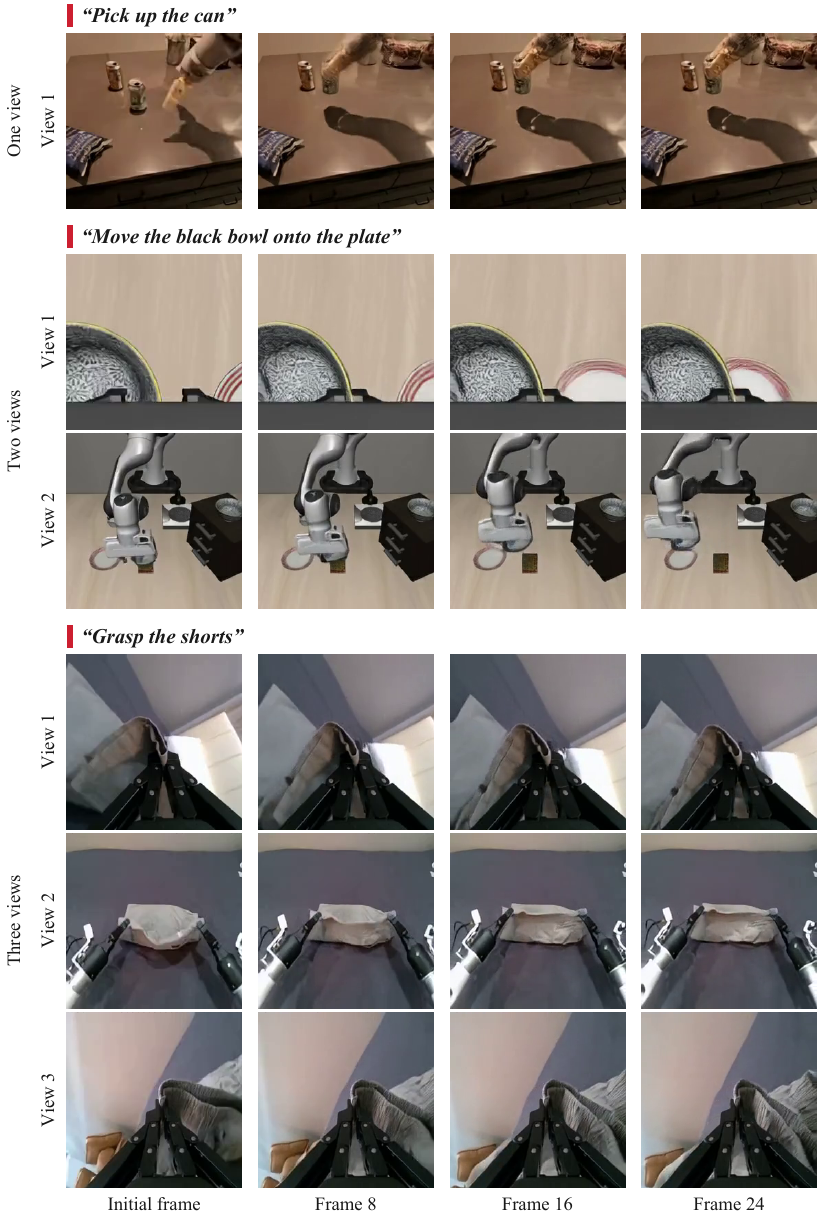}\par\nointerlineskip
  \caption{\textbf{Future-video predictions after Stage 1 adaptation.} The three examples use one, two, and three input camera views. Within each example, the first column contains observed frames and the remaining columns show predictions at frames 8, 16, and 24 of the 24-frame horizon; multiple rows correspond to synchronized views of the same sequence.}\label{fig:supp_video_generation}
\end{figure}

\paragraph{Video adaptation.}
Figure~\ref{fig:supp_video_generation} shows one-, two-, and three-view predictions after Stage 1 adaptation.
The first column is the observed frame; subsequent columns show predicted future frames 8, 16, and 24 of the $K=24$ clip.
Rows within each example are synchronized camera views.

\begin{figure}[t]
  \centering
  \setlength{\parskip}{0pt}
  \setlength{\abovecaptionskip}{8pt}
  \includegraphics[width=\linewidth]{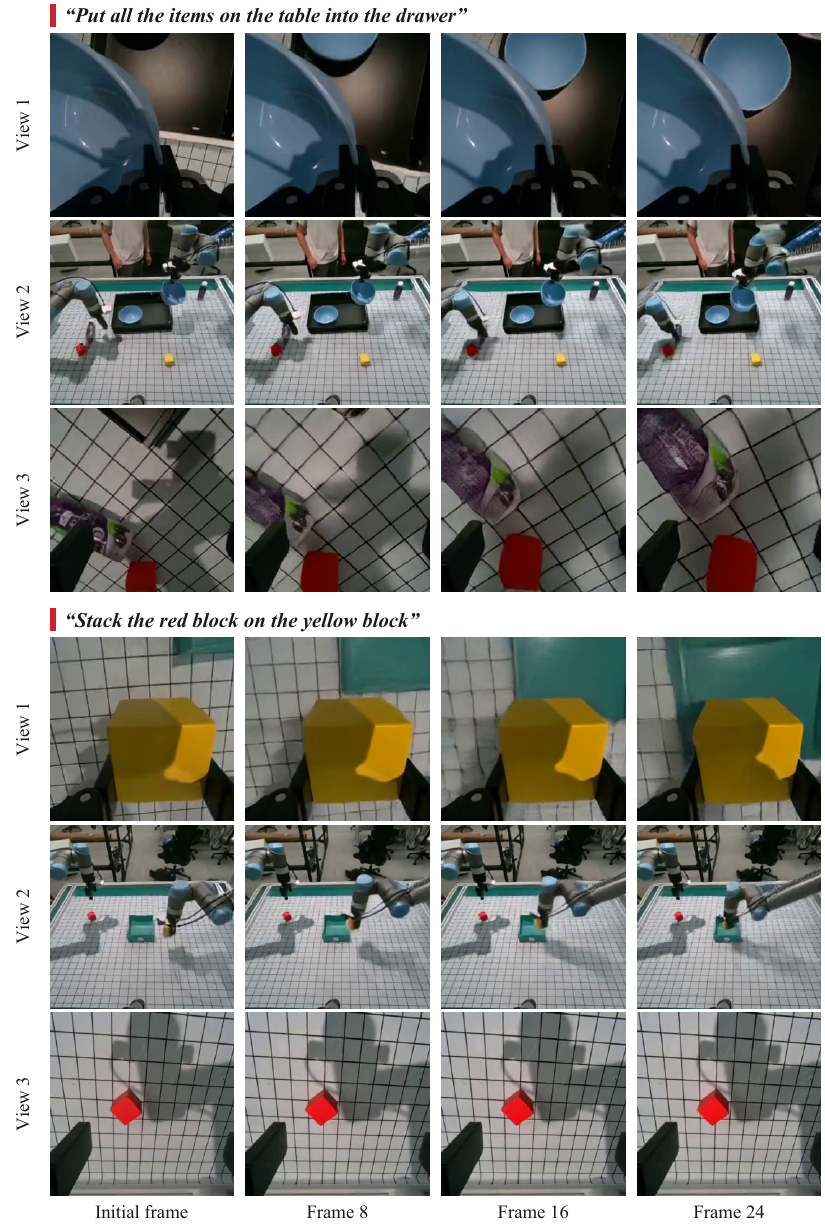}\par\nointerlineskip
  \caption{\textbf{Future-video predictions after real-world post-training.} Table clearing (top) and block stacking (bottom) each show three synchronized camera views. Observed frames occupy the first column, followed by predictions at frames 8, 16, and 24, so each row traces the predicted evolution of one camera view.}\label{fig:supp_realworld_video_generation}
\end{figure}

\paragraph{Real-world post-training.}
Figure~\ref{fig:supp_realworld_video_generation} shows three-view predictions after real-world post-training for table clearing and block stacking.
The temporal and camera conventions match Fig.~\ref{fig:supp_video_generation}.
These examples illustrate video prediction appearance and do not establish its quantitative relationship to control success.

\section{Supplementary Videos}
\label{sec:supp-videos}

The supplementary package contains six real-world videos corresponding to the six tasks described in \mainsectionref{sec:supp-realworld}.
The policy uses synchronized three-view observations; the videos display the third-person overview.
Representative image sequences are included in \mainfigureref{fig:realworld_vis}.

\begin{itemize}
    \item \texttt{01\_stack\_bowl\_and\_block.mp4}: Stack bowl and block.
    \item \texttt{02\_put\_bottle\_into\_trash\_can.mp4}: Put bottle into trash can.
    \item \texttt{03\_pour\_water\_from\_teapot.mp4}: Pour water from teapot.
    \item \texttt{04\_move\_bowl\_into\_tray.mp4}: Move bowl into a tray.
    \item \texttt{05\_stack\_two\_blocks.mp4}: Stack two blocks.
    \item \texttt{06\_clear\_the\_table.mp4}: Clear the table.
\end{itemize}

The videos are qualitative examples and are not used to compute the reported success rates.

\end{document}